# KNOWLEDGE AND UNCERTAINTY

Henry E. Kyburg, Jr.

One purpose -- quite a few thinkers would say the main purpose -- of seeking knowledge about the world is to enhance our ability to make good decisions. An item of knowledge that can make no conceivable difference with regard to anything we might do would strike many as frivolous. Whether or not we want to be philosophical pragmatists in this strong sense with regard to everything we might want to enquire about, it seems a perfectly appropriate attitude to adopt toward artificial knowledge systems.

If it is granted that we are ultimately concerned with decisions, then some constraints are imposed on our measures of uncertainty at the level of decision making. If our measure of uncertainty is real valued, then it isn't hard to show that it must satisfy the classical probability axioms. For example, if an act has a real-valued utility $U(E)$ if event $E$ obtains, and the same real-valued utility if the denial of $E$ obtains, ($U(E) = U(-E)$) then the expected utility of that act must be $U(E)$, and that must be the same as $p*U(E) + q*U(-E)$, where $p$ and $q$ represent the uncertainty of $E$ and $-E$ respectively. But then we must have $p + q = 1$.

1. There are reasons for rejecting real-valued -- i.e., strictly probabilistic -- measures of uncertainty, though not all the reasons that have been adduced for doing so are cogent. One is that these probabilities seem to embody more knowledge than they should: for example, if your beliefs are probabilistic, and you assign a probability of .01 to a drawn ball's being purple (on no evidence), and a probability of .02 to a second ball's being purple on the evidence that the first one is, and regard pairs of balls as "exchangeable", then you should be 99% sure that in the infinitely long run, no more than 11% of the balls will be purple. You know beyond a shadow of a doubt (with probability .99996) that no more than half will be purple. (In fact, we need much less than full exchangeability for this: all we need is that both individual events and pairs of events are treated the same way -- Kyburg, 1968.)

Peter Cheeseman (1985) has given a defense of classical probability, and perhaps would not find even such results as the foregoing distasteful. But it is hard to see how to defend the real-valued point of view from charges of subjectivity. Cheeseman refers to an "ideal" observer, but offers us no guidance in how to approach ideality, nor any characterization of how the ideal observer differs from the rest of us. It is therefore quite unclear what the ideal observer offers us, other than moral support: each of us is no doubt convinced that the ideal observer assigns probabilities just like himself. One man's subjective bias is another man's rational insight.

One defense against charges of subjectivity is to be found in the "convergence" theorems, of which the most famous is de Finetti's (de Finetti, 1937). Roughly: If $S$ is a sequence of trials, each resulting in success or failure, and you and I agree that the sequence is exchangeable, then no matter how divergent our initial views of the probability of

151

success -- so long as they are not given by probabilities of 0 or 1 -- and no matter what degree of agreement we seek, there is a number $\underline{n}$ such that after observing $\underline{n}$ trials you and I will agree to that degree on the probability of success on the next trial.

Of course each such theorem has a less gratifying counterpart: given any degree of disagreement that is intolerable, and given any $\underline{n}$, there exists a degree of initial disagreement such that <u>even</u> after $\underline{n}$ trials our degree of disagreement about the probability of success on the $\underline{n}$ plus first trial will be intolerable. And nothing precludes our disagreeing that much to start with.

It could be argued -- and has been -- that subjective probabilities don't vary so dreadfully much, and so in fact subjectivity is a mere hobgoblin. It may be <u>philosophically</u> troubling to those of that turn of mind, but it is of little practical importance.

But I think it can be argued that a small difference in some subjective probabilities can lead to a very large difference in others. Furthermore, it is well known that we all fail to conform to the probability calculus in our degrees of belief. That just means that we have to make some adjustments. Since small differences can lead to large ones, the particular adjustments we make can have large consequences.

There are other ways of representing uncertainty than by real numbers between 0 and 1. If these uncertainties are to be used in making decisions, however, they must be <u>compatible</u> with classical point-valued probabilities. My preference is for <u>intervals</u>, because they can be based on objective knowledge of distributions, and because this compatibility is demonstrable. (Kyburg, 1974)

In what follows, I will sketch the properties of interval-valued epistemic probability, and exhibit a structure for knowledge representation that allows for both uncertain inference from evidence and uncertain knowledge as a basis for decision. Along the way I make some comparisons to other approaches.

2. <u>Probability</u>.

Probability is a function from <u>statements</u> and <u>sets of statements</u> to closed subintervals of $[0,1]$. The sets of statements represent hypothetical bodies of knowledge. The idea behind $\underline{Prob}(\underline{S},\underline{K}) = [\underline{p},\underline{q}]$ is that someone whose body of knowledge is $\underline{K}$ should, <u>ought</u> to, have a 'degree' of belief in $\underline{S}$ characterized by the interval $[\underline{p},\underline{q}]$. The cash value of having such a 'degree' of belief is that he should not sell a ticket that returns to the purchaser \$1.00 for less than 100$\underline{p}$ cents, and he should not buy such a ticket for more than 100$\underline{q}$ cents. The relation in question is construed as a purely objective, logical relation.

Every probability can be based on <u>knowledge</u> of statistical distributions or relative frequencies, since statements known to have the same truth value receive the same probability, and every such equivalence class of statements (we can show) contains some statements of the appropriate form. This statistical knowledge may be both uncertain and approximate (we may be practically sure that between 30% and 40% of the

152

balls are black), but it is objective in the sense that any two people having the same evidence _should_ have the same knowledge.

Classical point-valued probabilities constitute a special case, corresponding to the extreme hypothetical (and unrealistic) case in which $\underline{K}$ embodies exact statistical knowledge.

The connection between statements and frequencies is given by a set of formal procedures for finding the right reference class for a given statement. The reference set may be multi-dimensional -- the set of urns, each paired with the set of draws made from it. It may be only "accidentally" related to sentence -- as when we predict the act of someone who makes a choice on the basis of a coin toss. What is the right reference class for a given statement $\underline{S}$ depends (formally and objectively) on what is in $\underline{K}$, our body of knowledge. In some cases we can implement a procedure for finding the right reference class.

It is natural to suppose that statistical knowledge in $\underline{K}$ is represented by the attribution to each reference set of a convex set of distributions -- for example we have every reason in the world to suppose that heads among coin-tosses in general is nearly binomial, with a parameter _close_ to a half. (We have no reason to suppose that the parameter has the real value .49999...). Or we may have good reason to believe that two quantities are uncorrelated in their joint distribution. Or that we can rule out certain classes of extreme distributions. We can know of a certain bent coin that heads will be binomially distributed in sequences of its tosses, with a parameter $\underline{p}$ at least equal to a half. In a wide range of cases of practical importance, what we can know of the set of distributions is that conditional independence obtains between certain variables. (Judea Pearl has made conditional independence the cornerstone of his constraint propagation approach (Pearl, 1985); conditional independence is what is required to warrant the use of Dempster's rule of combination.

Henceforth, we assume convexity. Here are some immediate results (Kyburg 1961, 1974):

(1) If $\underline{Prob}(\underline{S},\underline{K}) = [\underline{p},\underline{q}]$ then $\underline{Prob}(-\underline{S},\underline{K}) = [1-\underline{q}, 1-\underline{p}]$.

(2) If $\sim(\underline{S}\ \&\ \underline{T})$ is in $\underline{K}$, and $\underline{Prob}(\underline{S},\underline{K}) = [\underline{p}1,\underline{q}1]$ and $\underline{Prob}(\underline{T},\underline{K}) = [\underline{p}2,\underline{q}2]$ and $\underline{Prob}(\underline{T} \vee \underline{S}) = [\underline{p},\underline{q}]$, then there are numbers in $[\underline{p}1,\underline{q}1]$ and $[\underline{p}2,\underline{q}2]$ whose sum is in $[\underline{p},\underline{q}]$. To see that $[\underline{p},\underline{q}]$ can be a _proper_ subset of $[\underline{p}1 + \underline{p}2, \underline{q}1 + \underline{q}2]$, consider a die that you know to be biassed toward the one at the expense of the two, or toward the two at the expense of the one. Reasonable probability for the disjunction, "one or two" would be very close to 1/3, even though the reasonable probabilities for "one" and "two" would be significantly spread above and below 1/6.

(3) We can show that: given any finite set of sentences, $\underline{Si}$, and a body of knowledge $\underline{K}$, there exists a Bayesian function $\underline{B}$, satisfying the classical probability axioms, such that for every sentence $\underline{S}$ in $\underline{Si}$, $\underline{B}(\underline{S}) \in \underline{Prob}(\underline{S},\ \underline{K})$.

(4) Let $\underline{KE}$ be the body of knowledge obtained from $\underline{K}$ when evidence $\underline{E}$ is added to $\underline{K}$. If $\underline{E}$ is among the finite set of sentences in question, then

153

there may be no Bayesian function $\underline{B}$ satisfying both $\underline{B}(\underline{S}) \in \underline{Prob}(\underline{S},\underline{K})$ and $\underline{B}(\underline{S}/\underline{E}) \in \underline{Prob}(\underline{S},\underline{KE})$: classical conditionalization is not the only way of updating probabilities.

(5) The randomness relation is definable, and in fact for one kind of database rules for picking the right reference class have been implemented.

3. <u>Uncertain Knowledge</u>.

One problem that Bayesian and other approaches to uncertainty have is that there is no formal way of representing the acquisition of knowledge. We can represent the <u>having</u> of knowledge (by the assignment of probability 1 to the item), but since there is no way in which $\underline{P}(\underline{S}/\underline{E})$ can be 1 unless $\underline{P}(\underline{S})$ is already one, conditionalization doesn't get us knowledge. This has been noticed, of course; Cheeseman (1985, p. 1008) simply says, "A reasonable compromise is to treat propositions whose probability is close to 0 or 1 as if they are known with certainty..." But of course it is well known that this cannot be done generally: the conjunction of a number of certainties is a certainty, but the conjunction of a large enough number of certainties in Cheeseman's sense is what he would have to consider an impossibility!

McCarthy and Hayes (1969) are seduced into following this primrose path, when they suggest (p. 489) "If $\theta 1, \theta 2, \ldots, \theta n \vdash \theta$ is a possible deduction, then $\underline{probably}(\theta 1), \ldots, \underline{probably}(\theta n) \vdash \underline{probably}(\theta)$ is also a possible deduction." This is clearly ruled out, on our scheme -- and even $\underline{acceptable}(\theta 1), \ldots, \underline{acceptable}(\theta 2) \vdash \underline{acceptable}(\theta)$ is ruled out as a consequent of the logical conditional. If we are to formalize uncertain inference at all and not merely the deductive propagation of probabilities, we must somehow accommodate sets of conflicting statements. <u>Purely</u> probabilistic rules of inference do this easily.

We can accommodate Cheeseman's intuition that we should accept what is "practically certain" by considering <u>two</u> sets of sentences in the representation of knowledge. One of them we will call the evidential corpus, and denote by $\underline{Ke}$; the other we will call the practical corpus, and denote by $\underline{Kp}$.

We will accept a sentence into $\underline{Kp}$ if and only if its lower probability relative to $\underline{Ke}$ is greater than $\underline{p}$. The conjunction of two statements that appear in $\underline{Kp}$ will also appear in $\underline{Kp}$ only if the conjunction itself is probable enough relative to $\underline{Ke}$. Thus $\underline{Kp}$ will <u>not</u> be deductively closed, though we can prove that if a statement $\underline{S}$ appears in $\underline{Kp}$, and $\underline{S}$ entails $\underline{T}$, $\underline{T}$ may also appear there because it will have a lower probability greater than that of $\underline{S}$. This reflects a natural feature of human inference: we must have reason, not only to accept each premise in a complex argument, but to accept the <u>conjunction</u> of the premises, in order to be confident of the conclusion.

In fact, the uncertain inference that generates $\underline{Kp}$ from $\underline{Ke}$ has a number of the desirable features of non-monotonic inference. Add "Tweety is a bird" to $\underline{Ke}$, and "Tweety is capable of flight" will appear in $\underline{Kp}$ exactly because practically all birds fly. In addition, add "Tweety is an ostrich" to $\underline{Ke}$, and "Tweety is <u>not</u> capable of flight" will appear in $\underline{Kp}$.

In the former case, you should base your decisions on the assumption that Tweety can fly; in the latter, you no longer need worry about that possibility.

But to warrant the <u>detachment</u> that yields the addition of a sentence to our stock of practical certainties, we need more than a mere preponderance of evidence. We don't want to infer that two tosses of a coin will yield one head and one tail just because this is the most likely outcome. Similarly, we don't want to infer that a die will not yield a six: we want to say that the <u>probability</u> of an outcome other than a six is about five sixths.

This is just to say that the <u>level of practical certainty</u> $p$ is exactly what distinguishes (in a given context) sentences that we are willing to bet against from sentences that we take for granted.

We have a picture that looks like this:

```
********
*      *
*  Ke  *
********
    ⇓          Uncertain inference:  S ∈ Kp iff
                   Prob(S,ke) ≥ p.
********
*      *
*  Kp  *
********
```

It is relative to $\underline{Kp}$, the practical corpus, that we make our (practical) decisions. It is thus the (convex sets of) distributions -- including conditional distributions -- embodied in the practical corpus that we use in our decision theory.

But there are questions. What is the value of $\underline{p}$ that we are taking as practical certainty? How do statements get in $\underline{Ke}$? What is the decision theory that goes with this kind of structure?

Let us first consider the value of $\underline{p}$. Suppose the widest range of stakes we can come up with is 99:1. For example, Sam and Sally are going to bet on some event, each has $100, and neither has any change. Then a probability value falling outside the range of $[.01,.99]$ would be useless as a betting guide. A probability less than .01 would (in this context) amount to a practical impossibility; one greater than .99 would amount to a practical certainty.

The range of stakes can determine the level of "practical certainty" $\underline{p}$. What counts as practical certainty depends on context, but in an explicit way: it depends on what's at stake.

How do statements qualify as evidence in $\underline{Ke}$? Not by being "certain." It can be argued that anything that was really incorrigible would have to be devoid of empirical content. (The worry about uncertain evidence is not misplaced; it's just misconstrued.) One typical form of evidence statement is this: "The length of $\underline{x}$ is $\underline{d} \pm \underline{r}$ meters". Whatever our

155

readings, these statements are not "certain" -- they admit of error. The same is true of all ordinary observation statements.

So a statement gets into $\underline{Ke}$ by having a low probability of being in error; equally, by having a high probability (at least $\underline{e}$) of being veridical. How high? In virtue of the fact that conjunctions of pairs of statements in $\underline{Ke}$ appear in $\underline{Kp}$, it seems plausible to take $\underline{e} = (\underline{p})^{1/2}$. For a number of technical reasons (Kyburg, 1984) it turns out to be best to construe the corpus containing the theory of error as metalinguistic. This is as one might think: after all, the theory of error concerns the relation between <u>readings</u> -- e.g. <u>numerals</u> written in laboratory books -- and <u>values</u>: the real quantities characterizing things in the real world. For present purposes we need note only that this is <u>not</u> the beginning of an infinite regress. We <u>can</u> maintain objectivity; we <u>can</u> avoid "presuppositions" and other unjustified assumptions.

4. <u>Decision</u>.

It has been objected (Seidenfeld, 1979) that there is no decision theory that is tailored to Shafer's theory of evidential support. Indeed, it is pretty clear that support functions alone would conflict with expected utility. On the other hand, since Shafer's system of support functions is a special case of the representation by convex sets of distributions, we can have very nearly a normal decision theory using Shafer's system. In computing the value of an act, we need to consider not only the <u>support</u> assigned to various states of affairs (corresponding to lower probabilities), but also the <u>plausibilities</u> -- corresponding to the upper probabilities.)

This is true for the more general convex set representation: We can construct an <u>interval</u> of expected utility for each act. A natural reinterpretation of the principle of dominance would take an alternative $\underline{a}1$ to dominate an alternative $\underline{a}2$ whenever, for every possible frequency distribution, the expectation of $\underline{a}1$ is greater than the expectation of $\underline{a}2$.

This eliminates some alternatives, but in general there will be a number of courses of action that are not eliminated. What we do here is another matter, one which is certainly worthy of further study. But it seems natural that minimax and minimax regret strategies are appropriate candidates for consideration under some conditions. There may well be others, such as satisficing. And it may even by that the guidance provided by the motto: eliminate dominated alternatives, is as far as rationality alone takes us. Further pruning may depend on constraints that are local to the individual decision problems.

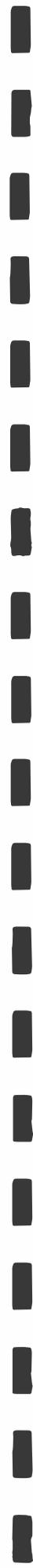